\documentclass[conference]{IEEEtran}
\IEEEoverridecommandlockouts
\usepackage{cite}
\usepackage{amsmath,amssymb}
\usepackage{graphicx}
\usepackage{booktabs}
\usepackage{multirow}
\usepackage{array}
\usepackage{xcolor}
\usepackage{etoolbox}
\usepackage[font=small,skip=0pt]{caption}
\usepackage{placeins}
\usepackage{url}
\usepackage[hidelinks]{hyperref}
\graphicspath{{Figures/}}
\usepackage{setspace}
\renewcommand\IEEEkeywordsname{Index Terms}

\newcommand{\vs}{VoxelSynth3D}

\newcommand{\rev}[1]{#1}
\def\BibTeX{{\rm B\kern-.05em{\sc i\kern-.025em b}\kern-.08em
T\kern-.1667em\lower.7ex\hbox{E}\kern-.125emX}}

\makeatletter
\let\oldthebibliography\thebibliography
\let\endoldthebibliography\endthebibliography
\renewenvironment{thebibliography}[1]{%
  \oldthebibliography{#1}%
  \fontsize{8}{7.3}\selectfont
  \setlength{\itemsep}{0pt}%
  \setlength{\parskip}{0pt}%
  \setlength{\parsep}{0pt}%
  \let\submittedbibitem\bibitem
  \renewcommand{\bibitem}[1]{%
    \ifstrequal{##1}{Haneda2025AAPM}{\color{black}}{%
      \ifstrequal{##1}{Peters2025Hybrid}{\color{black}}{%
        \ifstrequal{##1}{Wu2022XCIST}{\color{black}}{\color{black}}}}%
    \submittedbibitem{##1}}%
}{\endoldthebibliography}
\makeatother

\begin{document}

\title{VoxelSynth3D: Interpretable Volumetric Image-Domain Metal Artifact Reduction with a Paired Synthetic CLINIC Benchmark}


\author{
\IEEEauthorblockN{
Amritesh Banerjee\IEEEauthorrefmark{1}, Abdul Basit\IEEEauthorrefmark{2}, Renil Renji Joseph\IEEEauthorrefmark{3}, Nouhaila Innan\IEEEauthorrefmark{2}, Muhammad Shafique\IEEEauthorrefmark{2}}
\IEEEauthorblockA{\IEEEauthorrefmark{1}\textit{University of Massachusetts at Amherst}, Amherst, USA}
\IEEEauthorblockA{\IEEEauthorrefmark{2}\textit{eBRAIN Lab, Division of Engineering, New York University Abu Dhabi (NYUAD)}, Abu Dhabi, UAE}
\IEEEauthorblockA{\IEEEauthorrefmark{3}\textit{Indian Institute of Technology Delhi Abu Dhabi}, Abu Dhabi, UAE}
\IEEEauthorblockA{amriteshbane@umass.edu, 
renilrenjijm@gmail.com, 
\{abdul.basit, nouhaila.innan, muhammad.shafique\}@nyu.edu}
\vspace{-20pt}
}

\maketitle

\begin{abstract}
Metal artifacts in postoperative musculoskeletal CT obscure bone--implant and adjacent soft-tissue interfaces. Many metal artifact reduction (MAR) methods require unavailable raw projections or learned models that may shift across scanners and implants. We present \vs, a training-free 3D image-domain framework for reconstructed CT. \rev{The framework combines} support masking, normalized tissue \rev{synthesis}, deviation gating, and restricted edge refinement. Detected implant voxels are preserved in the output, while correction targets metal-induced artifacts in the surrounding tissue. We also construct \rev{Synthetic CLINIC-Metal, a controlled paired synthetic evaluation resource,} from no-metal CTPelvic1K volumes with clean targets, metal/artifact masks, fixed seeds, and patient-level splits; 75 unpaired real metal cases receive qualitative/no-reference evaluation only. The operating point was fixed in a near-flat validation basin. \rev{With exact-mask oracle localization, all methods share a metal-excluded tissue ROI. On 40 held-out cases, \vs{} reduced RMSE from 801.48 to 786.18 HU (paired gain 15.30 HU, 95\% CI 11.68--19.23), improving every case and exceeding the evaluated 3D Gaussian smoother by 13.58 HU. Clean-edge agreement decreased next to metal but exceeded input beyond 5 mm. Thus, \vs{} provides case-consistent within-distribution tissue-error reduction with a localized structural tradeoff. Spacing-aware sensitivity retained aggregate broad-region improvement and identified near-metal calibration as a target.}
\end{abstract}


\section{Introduction}
Computed tomography (CT) is indispensable for postoperative musculoskeletal assessment because it visualizes implant placement, osseointegration, loosening, infection, fracture fixation, and periprosthetic complications. The same hardware also corrupts CT: beam hardening, scatter, and photon starvation produce streaks and dark bands that obscure bone--implant and soft-tissue interfaces \cite{Wellenberg2018,Meyer2010}. As orthopedic hardware becomes more prevalent \cite{Misir2025,Goriainov2014}, metal artifact reduction (MAR) is a practical requirement for reliable interpretation.

Classical projection-domain methods, including linear interpolation (LI) and normalized MAR (NMAR), are deterministic and physically motivated, but their strongest form assumes raw or scanner-consistent projection data \cite{Kalender1987,Meyer2010,Gjesteby2016}. This assumption is restrictive for retrospective studies, public CT repositories, and multi-institutional datasets where only reconstructed volumes are available. Learning-based MAR, including image-domain CNNs and dual-domain models such as DuDoNet and InDuDoNet, can model nonlinear artifacts when training and deployment conditions match \cite{ZhangYu2018,Lin2019,Wang2021InDuDoNet,Wang2021InDuDoNetPlus}. However, these methods depend on training data, acquisition geometry, reconstruction kernels, and synthetic-to-clinical transfer. Reconstructed-volume CT therefore needs MAR methods that are volumetric, localized, interpretable, and evaluable without unavailable raw projections or opaque learned behavior.

\vs{} addresses this gap as a deterministic 3D image-domain framework. Correction is restricted to an implant-adaptive artifact support, where a normalized volumetric tissue prior estimates plausible non-metal attenuation, a deviation gate limits replacement to corrupted voxels, and restricted edge-guided refinement reinjects high-frequency structure only where support and gradients indicate softened anatomy. This design targets the central MAR tradeoff: suppressing streak and dark-band energy while limiting changes to unaffected tissue. \rev{Detected implant voxels are preserved in the output, while correction targets metal-induced artifacts in the surrounding tissue. The novelty is system-level: established operators are composed in an implant-adaptive 3D correction whose support, prior, and weights are inspectable, rather than proposed as new filtering primitives.}

\begin{figure*}[ht]
\centering
\includegraphics[width=\textwidth]{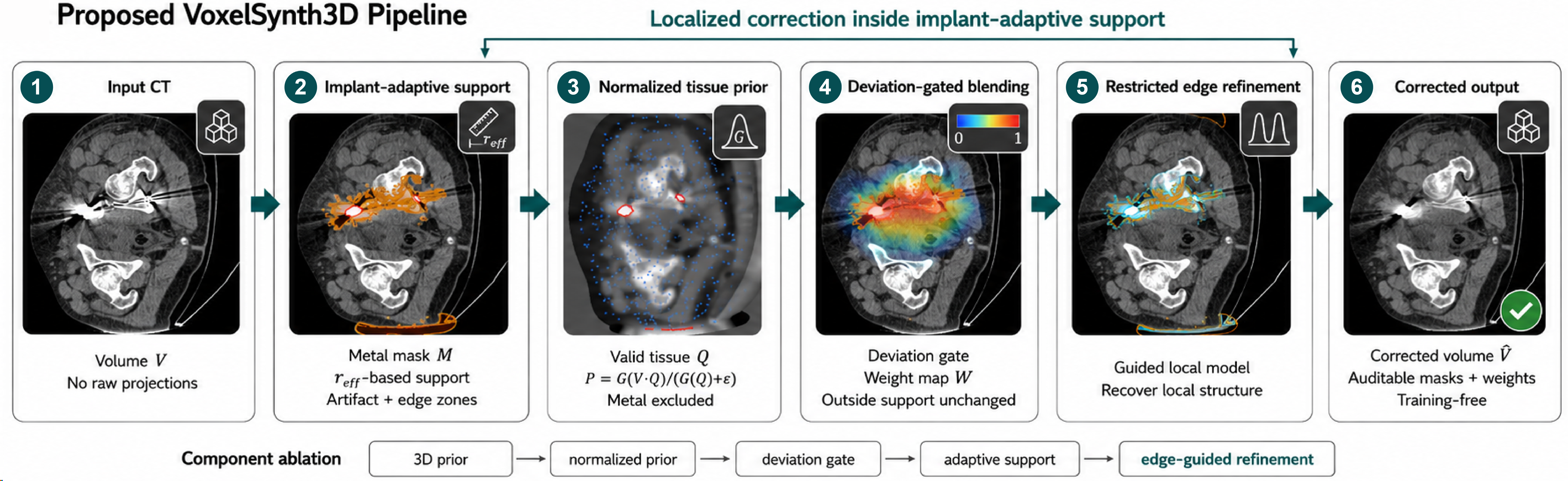}
\caption{ \footnotesize Workflow overview and visual intuition for the proposed \vs{} pipeline. The method operates directly on reconstructed CT volumes without raw projection data, estimates an implant-adaptive support, synthesizes a normalized non-metal tissue prior, applies deviation-gated local blending, and restricts edge-guided refinement to the artifact support. The component ablation path reports how each design element contributes to the final edge-guided configuration. \rev{Detected metal is retained in the reported output.}}
\label{fig:method_overview}
\end{figure*}

Evaluation is also difficult because real metal CT rarely has paired clean reconstructions. We therefore construct \rev{\textbf{Synthetic CLINIC-Metal}, a controlled paired synthetic evaluation resource. This synthetic dataset is generated} from no-metal CTPelvic1K pelvic CT volumes \cite{Liu2020CTPelvic1K}, retaining clean targets, corrupted volumes, metal masks, artifact-ROI masks, spacing metadata, and patient-level splits. This enables \rev{within-distribution} full-reference evaluation on synthetic cases while keeping real CLINIC-metal scans separate for qualitative and no-reference validation. \rev{Full-reference evaluation uses one fixed surrounding-tissue ROI after excluding the inserted core.} Pseudo-sinogram baselines are included for comparison, but are interpreted as approximate image-domain adaptations rather than scanner-native projection MAR. \rev{Learned rows are likewise public-checkpoint transfer tests, not controlled in-domain comparisons.}

This paper makes four contributions:
\begin{itemize}
    \item \textbf{Volumetric MAR:} adaptive support, normalized tissue synthesis, gated blending, and restricted edge refinement for reconstructed CT.
    \item \textbf{\rev{Controlled paired resource:}} Synthetic CLINIC-Metal with clean targets, masks, spacing, patient-level splits, \rev{and a metal-excluded tissue endpoint}.
    \item \textbf{Comparative evaluation:} image-domain filters, pseudo LI/NMAR, learned MAR \rev{checkpoint-transfer tests}, and component ablations.
    \item \textbf{Conservative protocol:} validation sensitivity, CIs, paired tests, severity, and real no-reference analysis\rev{, plus distance, edge, mask, and external stress tests}.
\end{itemize}

On 40 held-out synthetic cases, \vs{} achieved lower \rev{metal-excluded tissue} RMSE and higher ROI MSE reduction than the \rev{input and evaluated 3D Gaussian smoother}, while learned baselines retained higher edge preservation or \rev{tissue SSIM} in some settings. \rev{The effect was case-consistent: every case improved, while clean-target analysis found a true-edge cost immediately next to metal. The claim is therefore within-distribution artifact suppression with a localized structural tradeoff, not anatomical recovery.} We do not claim scanner-native projection-domain superiority; instead, we target retrospective settings where only reconstructed CT is available. Figure~\ref{fig:method_overview} illustrates the VoxelSynth3D pipeline.

\section{Related Work}
\subsection{Classical and Projection-Domain MAR}
Metal artifacts arise because high-density objects violate assumptions used by standard CT reconstruction. Linear interpolation replaces corrupted metal traces before filtered backprojection, while NMAR normalizes the sinogram with a prior image before interpolation to reduce discontinuities \cite{Meyer2010}. These methods remain clinically relevant because they are simple and interpretable, but they can erase structures crossing the metal trace and introduce new streaks. Their strongest form requires access to scanner projection data; when only reconstructed images are available, pseudo-sinogram baselines become approximate and should be interpreted cautiously.

\subsection{Learning-Based and Hybrid MAR}
Learning-based MAR has progressed from image-domain convolutional networks to dual-domain approaches that jointly process sinograms and images. DuDoNet introduced a dual-domain formulation for CT MAR \cite{Lin2019}; InDuDoNet and InDuDoNet+ further unfolded interpretable dual-domain iterations \cite{Wang2021InDuDoNet,Wang2021InDuDoNetPlus}. Dictionary-style unrolled models such as ACDNet also use learned sparse priors for MAR \cite{Wang2022ACDNet}. These models can be powerful when the deployment domain matches the training distribution, but supervised learning requires paired or synthetic data and may be sensitive to scanner geometry, implant appearance, and reconstruction kernels. Explainability concerns in medical AI \cite{Tjoa2021} and robustness concerns under distribution shift \cite{Goodfellow2015,Zhang2023} motivate methods whose correction support, blending weights, and anatomical priors are inspectable. \vs{} follows this direction while remaining fully image-domain and volumetric.

Commercial scanner MAR and orthopedic MAR workflows are clinically useful when available \cite{Wellenberg2018}, but reconstructed-image methods remain relevant for retrospective archives, public datasets, and multi-site studies where raw projections, vendor reconstructions, and kernel details are unavailable.

\rev{The 2025 AAPM CT-MAR Grand Challenge provides a standardized 2D multi-anatomy benchmark, while its hybrid database combines clinical images, virtual metals, and XCIST/CatSim physics \cite{Haneda2025AAPM,Peters2025Hybrid,Wu2022XCIST}. Synthetic CLINIC-Metal is complementary rather than a replacement: it supplies paired 3D pelvis volumes and spatial masks for localization analysis.}

Table~\ref{tab:soa_positioning} summarizes this positioning across methodological assumptions and evaluation evidence.

\begin{table}[ht]
\centering
\caption{Compact positioning of representative MAR \rev{methods and their reported evaluation evidence}.}
\label{tab:soa_positioning}
\scriptsize
\setlength{\tabcolsep}{1.8pt}
\renewcommand{\arraystretch}{1.10}
\newcommand{\cmark}{\textcolor{green!50!black}{$\checkmark$}}
\newcommand{\xmark}{\textcolor{red!70!black}{$\times$}}
\newcommand{\pmark}{\textcolor{orange!85!black}{$\triangle$}}
\begin{tabular}{p{0.30\columnwidth}>{\centering\arraybackslash}p{0.095\columnwidth}>{\centering\arraybackslash}p{0.095\columnwidth}>{\centering\arraybackslash}p{0.085\columnwidth}>{\centering\arraybackslash}p{0.095\columnwidth}>{\centering\arraybackslash}p{0.105\columnwidth}>{\centering\arraybackslash}p{0.105\columnwidth}}
\toprule
\rev{Method} & \shortstack{Proj.\\phys.} & \shortstack{CT\\only} & \shortstack{3D\\vol.} & \shortstack{No\\train} & \shortstack{Paired\\GT} & \shortstack{\rev{Real}\\\rev{eval.}} \\
\midrule
LI/NMAR \cite{Meyer2010} & \cmark & \xmark & \xmark & \cmark & \pmark & \pmark \\
Pseudo LI/NMAR & \pmark & \cmark & \xmark & \cmark & \cmark & \pmark \\
DuDoNet \cite{Lin2019} & \cmark & \pmark & \xmark & \xmark & \pmark & \pmark \\
InDuDoNet/+ \cite{Wang2021InDuDoNet,Wang2021InDuDoNetPlus} & \cmark & \pmark & \xmark & \xmark & \pmark & \pmark \\
ACDNet family \cite{Wang2022ACDNet} & \pmark & \pmark & \xmark & \xmark & \pmark & \pmark \\
\textbf{\vs{} (ours)} & \xmark & \cmark & \cmark & \cmark & \cmark & \pmark \\
\bottomrule
\end{tabular}

\begin{minipage}{0.98\columnwidth}
\vspace{3pt}
\scriptsize \textit{Abbreviations:} Proj. phys.=projection/scanner physics; CT only=reconstructed CT without raw projections; 3D vol.=volumetric; No train=training-free; Paired GT=paired-reference evaluation; \rev{Real eval.=real-metal CT; \vs{} uses qualitative/no-reference evidence.} \cmark{} native, \pmark{} partial or implementation-dependent, \xmark{} absent.
\end{minipage}
\end{table}

\section{Methodology}

Let $V\in\mathbb{R}^{Z\times Y\times X}$ be the reconstructed CT volume in HU and let $\mathbf{s}=(s_z,s_y,s_x)$ denote voxel spacing. \vs{} produces a corrected volume $\hat V$ through four deterministic stages: adaptive support estimation, normalized tissue prior synthesis, deviation-gated blending, and restricted edge-guided refinement. This edge-guided configuration is the proposed method used in the main experiments; earlier components and later residual or external-cleanup variants are evaluated only as descriptive ablations. The method is applied in 3D so that correction weights vary smoothly across adjacent axial slices.
\begin{figure}[htbp]
\centering
\includegraphics[width=\columnwidth]{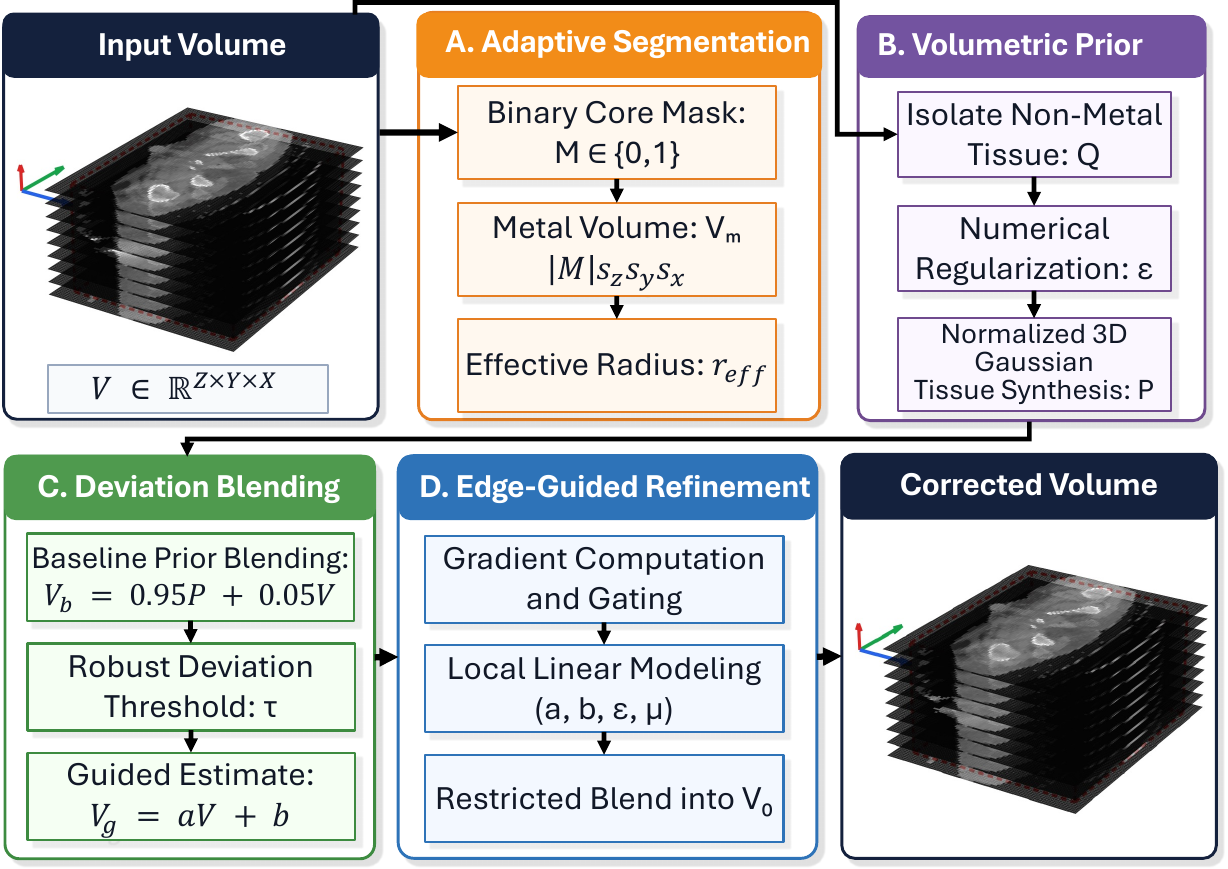}
\caption{\footnotesize \textbf{VoxelSynth3D methodology:} The corrupted reconstructed CT volume is processed through four deterministic stages: adaptive metal-support estimation, normalized volumetric tissue-prior synthesis, deviation-gated blending, and restricted edge-guided refinement. Together, these steps localize correction to artifact-corrupted regions, reinject supported high-frequency structure, and avoid global smoothing of unaffected anatomy.}
\label{fig:methodology_pipeline}
\end{figure}

\begin{table}[ht]
\centering
\caption{Default deterministic parameters used for the proposed configuration.}
\label{tab:default_params}
\scriptsize
\setlength{\tabcolsep}{2pt}
\renewcommand{\arraystretch}{1.02}
\begin{tabular}{p{0.34\columnwidth}p{0.58\columnwidth}}
\toprule
Component & Default value \\
\midrule
Metal seed & 99.5th percentile of voxels above 400 HU, clipped to 1000--1500 HU \\
Support & $r_m=\mathrm{clip}(1.6r_{\rm eff},8,22)$ mm; $r_e=\mathrm{clip}(0.8r_{\rm eff},4,10)$ mm \\
Valid tissue $Q$ & finite non-metal soft tissue, $-300<V<500$ HU; fallback: finite non-metal voxels \\
Prior/gate & normalized Gaussian prior, $\sigma=4$ mm; $\tau=\mathrm{clip}(2.5\sigma_Q,25,75)$ HU \\
Weight $W$ & 2 mm smoothed macro support, deviation gate, 0.70 Sobel-gradient attenuation \\
Blend/refinement & $V_b=0.95P+0.05V$; edge percentile 85; guided $\sigma=1$ mm; $\alpha=0.45$ \\
\bottomrule
\end{tabular}
\end{table}

\subsection{Adaptive Metal and Artifact Support}
For synthetic cases, $M$ is the saved inserted-metal mask\rev{, defining exact-mask oracle localization}; for real reconstructed CT, $M$ is seeded from an adaptive upper-tail HU threshold (99.5th percentile above 400 HU, clipped to 1000--1500 HU) and cleaned with 3D connected-component constraints. \rev{Boundary perturbations and automatic HU localization are evaluated separately from correction-stage performance.} The physical metal volume is $V_m = |M|\,s_zs_ys_x$,
which defines an effective implant radius $r_{\mathrm{eff}} = \left(\frac{3V_m}{4\pi}\right)^{1/3}$.
This radius scales the local correction support rather than using one global kernel for all implant sizes. A broader macro support captures dark-band and streak influence near the implant, while a narrower edge support limits high-frequency refinement to regions where anatomical edge recovery is plausible. \rev{Table~\ref{tab:default_params} summarizes the selected settings.}

\subsection{Normalized Volumetric Tissue Prior}
Naive Gaussian smoothing across metal voxels can contaminate the estimate with extreme HU values. We therefore compute a normalized tissue prior from valid non-metal tissue $Q$:
\begin{equation}
    P = \frac{G_{\sigma}(V\odot Q)}{G_{\sigma}(Q)+\epsilon},
\end{equation}
where $G_{\sigma}$ is an anisotropic Gaussian operator parameterized in millimeters and $\epsilon$ avoids division by zero. We use a denominator guard of $\epsilon=10^{-4}$ with a median-tissue fallback when local valid support is insufficient. The valid mask $Q$ excludes metal, nonfinite voxels, and non-soft-tissue extremes when enough samples are available. This normalized form estimates local tissue intensity from nearby valid voxels, \rev{stabilizing the prior around excluded metal}.

\subsection{Deviation-Gated Blending}
The prior is blended with the input only where the local intensity deviates from expected tissue statistics. A robust threshold $\tau$ is estimated from non-metal tissue variation and clipped to avoid overcorrection in low-noise or high-noise volumes. The baseline corrected image is
\begin{equation}
    V_b = 0.95P + 0.05V,
\end{equation}
followed by spatially weighted blending
\begin{equation}
    V_0 = W V_b + (1-W)V.
\end{equation}
The weight $W$ is nonzero only inside the adaptive artifact support, decays with distance from the implant, and is attenuated around strong uncorrupted tissue gradients. Specifically, the normalized Sobel gradient $g\in[0,1]$ reduces correction strength by $W{\leftarrow}W(1-0.70g)$ outside the metal core. Voxels outside the support remain unchanged, which is important for clinical interpretability and for preventing global smoothing.

\subsection{Edge-Guided Refinement}
The edge-guided refinement step addresses the tendency of volumetric smoothing to soften anatomical boundaries. Gradients are computed from the original input, and only high-gradient voxels within the correction support are eligible for edge reinjection. A local linear guided model is estimated between the input $V$ and baseline correction $V_0$:
\begin{equation}
    a = \frac{\operatorname{cov}(V,V_0)}{\operatorname{var}(V)+\epsilon}, \qquad
    b = \mu_{V_0}-a\mu_V,
\end{equation}
with local means and covariances computed by Gaussian windows. The guided estimate $V_g=aV+b$ is blended into $V_0$ with a conservative blend of 0.45 supported by validation sensitivity analysis and clipped to the valid CT range. This step is deliberately restricted: it \rev{partially mitigates local softening} within the artifact support, but it does not modify distant anatomy. \rev{Because attenuation inside metal is not estimated, the final output uses the parameter-free composition}
{
\begin{equation}
    \hat V=(1-M)\odot V_g+M\odot V,
\end{equation}
}
\rev{which leaves every detected metal voxel unchanged.} A multi-scale residual-matching extension is evaluated only as an ablation because it improves edge metrics at the cost of runtime and slightly worse primary RMSE.

\section{Experimental Setup}
\begin{figure*}[ht]
\centering
\includegraphics[width=\textwidth]{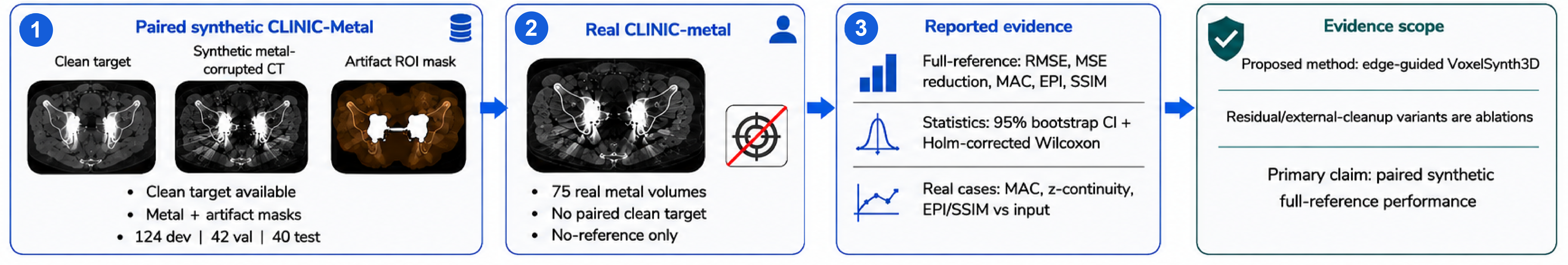}
\caption{Evaluation design. Paired synthetic CLINIC-Metal supplies clean targets and implant/artifact masks; real CLINIC-Metal has no paired target and supports only qualitative/no-reference analysis. \rev{The primary endpoint excludes the inserted metal core.}}
\label{fig:experimental_setup}
\end{figure*}

\subsection{\texorpdfstring{\rev{Paired Controlled Synthetic Evaluation Resource}}{Paired Controlled Synthetic Evaluation Resource}}
We constructed \rev{Synthetic CLINIC-Metal, a controlled paired synthetic evaluation resource,} from no-metal CLINIC pelvic CT volumes in CTPelvic1K \cite{Liu2020CTPelvic1K}; no-metal cases support synthesis and real-metal cases remain separate. \rev{It supports reproducible within-distribution benchmarking, not calibrated scanner physics or replacement of AAPM.} Clean CT was retained as ground truth; fixed-seed variants sampled 1--3 ellipsoidal implants from pelvis/body masks with size probabilities 0.50/0.35/0.15, random orientation and slice drift, metal attenuation $\mu=2.2$--$5.0~\mathrm{cm}^{-1}$, and 3071-HU metal voxels. Corruption used HU-to-attenuation conversion ($\mu_\mathrm{water}=0.192~\mathrm{cm}^{-1}$), 640-view Radon projection/filtered backprojection, metal-trace amplification, beam-hardening-like terms (0.02--0.20), and Poisson counts $I_0=1.5{\times}10^5$--$5{\times}10^5$. Artifact masks combined $|V^c-V^\ast|>30$ HU with an eight-voxel metal dilation inside the body. Source-patient splits were assigned before insertion: 124 development, 42 validation, and 40 test cases.

\subsection{\rev{Artifact-Model Sensitivity and Scope}}
\rev{To assess robustness beyond the primary controlled distribution, we evaluated the fixed method with a spacing-aware artifact model that incorporates pixel size when converting Radon sums to Beer--Lambert integrals. The evaluation used the first synthetic variant (index 00) from each of 21 validation patients, with criteria defined during the initial validation-only pilot and then held fixed: $\geq5\%$ broad-RMSE improvement, $\leq5\%$ interface worsening, local-SSIM loss $\leq0.01$, and $\geq75\%$ of cases within a 10\% interface margin. Importantly, broad-region error reduction remained evident under this changed artifact model: the mean method/input RMSE ratio was 0.891, with improvement in 13/21 cases. The sensitivity was concentrated near the metal interface, where the 2--5-mm RMSE ratio was 2.772 and local SSIM changed by $-0.150$. Excluding two cases under the fixed QC rule produced similar results (0.889/2.726; 0/19 within the interface margin). This analysis therefore shows that the broad-region correction effect is retained on average under altered artifact physics while identifying the immediate peri-metal region as the principal target for further physics calibration and refinement.}

\subsection{Real CLINIC-Metal Validation}
The real CLINIC-Metal subset contains 75 volumes with metallic implants but no paired clean reconstructions. We therefore use it only for qualitative and no-reference evaluation. Metrics on these cases measure how much a method changes artifact-like regions and how smooth the correction is across slices; they do not establish full-reference image quality.

\subsection{Comparators and Ablations}
Synthetic full-reference evaluation includes the uncorrected input, \rev{a descriptive slice-wise 2D Gaussian control,} 3D Gaussian filtering, pseudo LI-MAR/NMAR, DICDNet, InDuDoNet, InDuDoNet+, an adaptive-prior \vs{} ablation, the proposed edge-guided \vs{} method, and a multi-scale residual-matching extension. Pseudo LI/NMAR use approximate pseudo-sinograms rather than scanner raw projections. DICDNet and InDuDoNet-family models are public pretrained checkpoints applied as available-domain stress-test comparators, so their results reflect deployment transfer rather than the maximum possible performance after in-domain retraining. \rev{We therefore label learned rows as public-checkpoint transfer tests.} CPU wall-clock times include method/wrapper execution and metric bookkeeping and are implementation timings, not a controlled accelerator benchmark. We report the numerical validation-minimum edge setting for transparency but do not treat it as proposed. Real no-reference evaluation includes 2D median, 3D Gaussian, and the three \vs{} configurations. \rev{All correction parameters remain fixed. Eq.~(5) adds parameter-free implant-core pass-through. Exact-mask oracle experiments isolate correction under accurate localization, whereas automatic HU localization assesses the current heuristic end-to-end pipeline.}

\subsection{Metrics and Statistics}
\rev{The primary synthetic endpoint is RMSE in one fixed artifact-tissue ROI after excluding the exact inserted core, $\Omega_t=\Omega\setminus M$:}
{
\begin{equation}
    \mathrm{RMSE}_{\Omega_t}=\sqrt{\frac{1}{|\Omega_t|}\sum_{i\in\Omega_t}(\hat V_i-V_i^{\ast})^2}.
\end{equation}
}
Full-reference \rev{residual-error variance reduction} in the metal-excluded tissue ROI is \rev{$1-\operatorname{Var}_{\Omega_t}(\hat V-V^{\ast})/\operatorname{Var}_{\Omega_t}(V^c-V^{\ast})$; positive values indicate lower error variance than input.} \rev{Every method and mask condition uses the same $\Omega_t$ and clean target; MSE reduction is relative to input. A secondary metal-containing analysis found that 41.2\% of aggregate squared-error reduction arose inside metal, although metal formed only 0.218\% of ROI voxels. Spatial analysis excludes 0--2 mm to reduce boundary, mask, and partial-volume ambiguity, then uses non-overlapping 2--5, 5--10, and 10--20 mm bands. For each case, one 90th-percentile threshold is pooled from clean-target in-plane Sobel magnitudes in $\Omega_t$ and applied unchanged to every prediction and band, without prediction-specific normalization. Every second active axial slice is sampled deterministically to reduce adjacent-slice redundancy and computation; identical indices are used for all methods. Precision and recall use a one-pixel tolerance (0.75--0.98 mm). Complete prediction misses receive zero F1 before case averaging. Mask tests cover the exact-mask oracle, 1/2-mm erosion/dilation, HU localization, and collateral change outside the fixed ROI.} We report \rev{95\% percentile confidence intervals from 5000 case-level bootstrap resamples and use paired two-sided Wilcoxon tests for method comparisons. Within each metric, Holm correction is applied across nine comparator contrasts: input, 3D Gaussian, pseudo LI-MAR, pseudo NMAR, DICDNet, InDuDoNet, InDuDoNet+, VoxelSynth3D-A, and VoxelSynth3D-MS. Cases are grouped by} severity based on metal/ROI burden. For unpaired real cases, input-relative MAC, EPI, SSIM, and inter-slice energy only \rev{describe correction behavior}.

\subsection{\rev{External Stress Test}}
\rev{To assess behavior on an independent reconstructed-image dataset, we applied \vs{} without parameter adjustment to all 29 public AAPM CT-MAR scoring images. Because the challenge data are 2D, each image was replicated across five isotropic slices to satisfy the volumetric input requirement, and the center slice of the resulting output was evaluated. Reference images and scoring masks were used only for evaluation and were not available during inference, mask construction, or adaptation. This experiment is therefore interpreted as an external reconstructed-image stress test rather than a native 3D or clinical validation.}

\section{Results and Discussion}
\rev{The evaluation addresses three questions: whether error reduction survives metal exclusion, how structural cost varies with distance from metal, and how mask extent affects both the target score and unaffected tissue. These precede external and real-data analyses.}

\subsection{Full-Reference Synthetic \rev{Evaluation}}
Table~\ref{tab:main_results} summarizes held-out performance. \rev{To evaluate correction of the surrounding tissue independently of implant-core handling, all image-quality metrics are reported on a common metal-excluded tissue ROI. This distinction is important because metal occupies only 0.218\% of the original ROI yet accounts for 41.2\% of \vs{}'s aggregate squared-error reduction. The metal-excluded endpoint therefore focuses on artifact correction in adjacent tissue; metal-containing RMSE/EPI provide complementary sensitivity measures.}

\begin{table*}[t]
\centering
\caption{Held-out synthetic results ($n=40$). \rev{Image metrics share the metal-excluded tissue ROI and are mean [95\% bootstrap CI]}}
\label{tab:main_results}
\scriptsize
\setlength{\tabcolsep}{2.3pt}
\renewcommand{\arraystretch}{0.96}
\begin{tabular}{@{}lcccccc@{}}
\toprule
Method & \rev{RMSE (HU) $\downarrow$} & \rev{MSE red. (\%) $\uparrow$} & \rev{Err.-var. red. $\uparrow$} & \rev{EPI $\uparrow$} & \rev{SSIM $\uparrow$} & Time (s) $\downarrow$ \\
\midrule
Uncorrected Input & \rev{801.48 [789.40,813.35]} & \rev{0.00 [0.00,0.00]} & \rev{0.000 [0.000,0.000]} & \rev{0.311 [0.290,0.335]} & \rev{0.239 [0.232,0.247]} & 1.66 [1.55,1.78] \\
2D Gaussian & \rev{800.14 [787.89,811.90]} & \rev{0.34 [0.29,0.39]} & \rev{$-0.010$ [$-0.013$,$-0.007$]} & \rev{0.330 [0.308,0.354]} & \rev{0.239 [0.232,0.246]} & 14.58 [14.24,14.95] \\
3D Gaussian & \rev{799.75 [787.38,811.92]} & \rev{0.43 [0.38,0.49]} & \rev{$-0.008$ [$-0.011$,$-0.004$]} & \rev{0.333 [0.310,0.356]} & \rev{0.239 [0.232,0.246]} & 16.52 [16.21,16.87] \\
Pseudo LI-MAR & \rev{801.46 [789.33,813.52]} & \rev{0.00 [0.00,0.01]} & \rev{0.000 [0.000,0.000]} & \rev{0.310 [0.289,0.333]} & \rev{0.240 [0.232,0.247]} & 97.40 [79.98,116.52] \\
Pseudo NMAR & \rev{801.48 [788.96,813.54]} & \rev{0.00 [$-0.01$,0.00]} & \rev{0.000 [$-0.001$,0.000]} & \rev{0.310 [0.289,0.333]} & \rev{0.240 [0.232,0.247]} & 96.61 [78.43,115.75] \\
DICDNet\rev{$^\ast$} & \rev{819.35 [807.03,831.40]} & \rev{$-4.52$ [$-4.60$,$-4.44$]} & \rev{0.029 [0.027,0.031]} & \rev{0.301 [0.278,0.325]} & \rev{0.273 [0.264,0.282]} & 65.86 [55.42,78.50] \\
InDuDoNet\rev{$^\ast$} & \rev{802.85 [790.59,814.86]} & \rev{$-0.34$ [$-0.62$,$-0.05$]} & \rev{\textbf{0.056 [0.052,0.060]}} & \rev{0.296 [0.274,0.320]} & \rev{\textbf{0.288 [0.280,0.298]}} & 18.92 [17.60,20.44] \\
InDuDoNet+\rev{$^\ast$} & \rev{817.01 [805.11,828.69]} & \rev{$-3.93$ [$-4.09$,$-3.77$]} & \rev{0.048 [0.043,0.053]} & \rev{0.292 [0.271,0.315]} & \rev{0.270 [0.262,0.278]} & 18.81 [17.56,20.20] \\
\midrule
VoxelSynth3D-A & \rev{786.29 [773.38,799.11]} & \rev{3.75 [2.88,4.68]} & \rev{$-0.053$ [$-0.072$,$-0.036$]} & \rev{0.336 [0.314,0.360]} & \rev{0.244 [0.236,0.252]} & 33.09 [32.43,33.79] \\
\textbf{VoxelSynth3D} & \rev{\textbf{786.18 [773.28,798.81]}} & \rev{\textbf{3.78 [2.94,4.76]}} & \rev{$-0.053$ [$-0.072$,$-0.035$]} & \rev{\textbf{0.338 [0.315,0.362]}} & \rev{0.244 [0.236,0.252]} & 53.12 [52.19,53.95] \\
VoxelSynth3D-MS & \rev{786.47 [773.61,799.47]} & \rev{3.71 [2.84,4.64]} & \rev{$-0.052$ [$-0.072$,$-0.035$]} & \rev{0.338 [0.315,0.362]} & \rev{0.243 [0.236,0.251]} & 101.46 [99.46,103.36] \\
\bottomrule
\end{tabular}
\vspace{1pt}
\begin{minipage}{0.98\textwidth}
\vspace{2pt}
\scriptsize \rev{MSE and residual-error variance reductions are relative to input. $^\ast$Public-checkpoint transfer test.} Full-ROI sensitivity values are reported in text.
\end{minipage}
\end{table*}

\begin{figure}[t]
\centering
\includegraphics[width=\columnwidth]{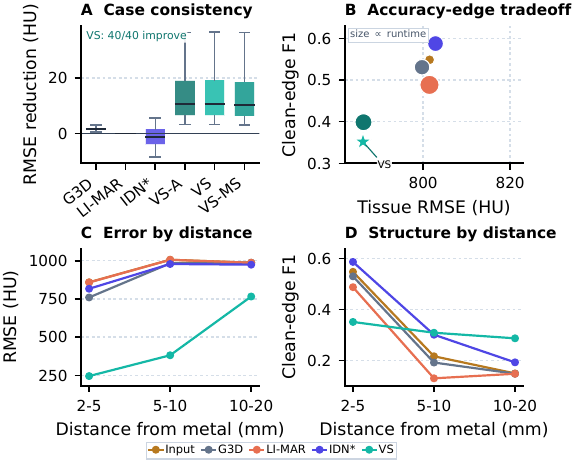}
\caption{\rev{Tissue-error and structure analysis. (A) Case-level RMSE reduction relative to input. (B) Tissue RMSE--clean-edge F1 tradeoff at 2--5 mm; marker size denotes runtime. (C--D) Distance-resolved RMSE and clean-target edge F1. IDN$^\ast$ denotes public-checkpoint transfer.}}
\label{fig:revised_evidence}
\end{figure}

\rev{On the metal-excluded tissue ROI, \vs{} achieved an RMSE of 786.18 HU, compared with 801.48 HU for the input and 799.75 HU for 3D Gaussian filtering. The corresponding paired gains were 15.30 HU (95\% CI 11.68--19.23) and 13.58 HU (10.07--17.40), respectively. The 1.9\% reduction was observed in all 40 held-out cases, with gains of 13.73, 17.21, and 12.66 HU across low-, medium-, and high-severity groups. These results show a consistent surrounding-tissue benefit that is not reproduced by Gaussian smoothing.}
\rev{Structural metrics provide complementary information. \vs{} achieved the highest tissue-ROI EPI (0.338), while InDuDoNet showed the highest SSIM and residual-error variance reduction. \vs{} had a residual-error variance reduction of $-0.053$. Because RMSE, variance reduction, and EPI capture different aspects of reconstruction error, they are interpreted jointly. In particular, EPI reflects gradient agreement over the broader tissue ROI, whereas clean-edge F1 evaluates target-defined anatomical edges by distance from metal. Complementary full-ROI RMSE/EPI values were 785.43/0.542 for \vs{}, 811.86/0.581 for input, 802.21/0.690 for pseudo LI-MAR, and 803.10/0.687 for InDuDoNet. Body SSIM was also stable relative to input (0.866 versus 0.865). Figure~\ref{fig:revised_evidence} shows case-consistent tissue-error gains across distance bands and exposes the localized edge tradeoff.}

\subsection{Ablation, Severity, and Metric Tradeoff}
\begin{figure}[h]
\centering
\includegraphics[width=\columnwidth]{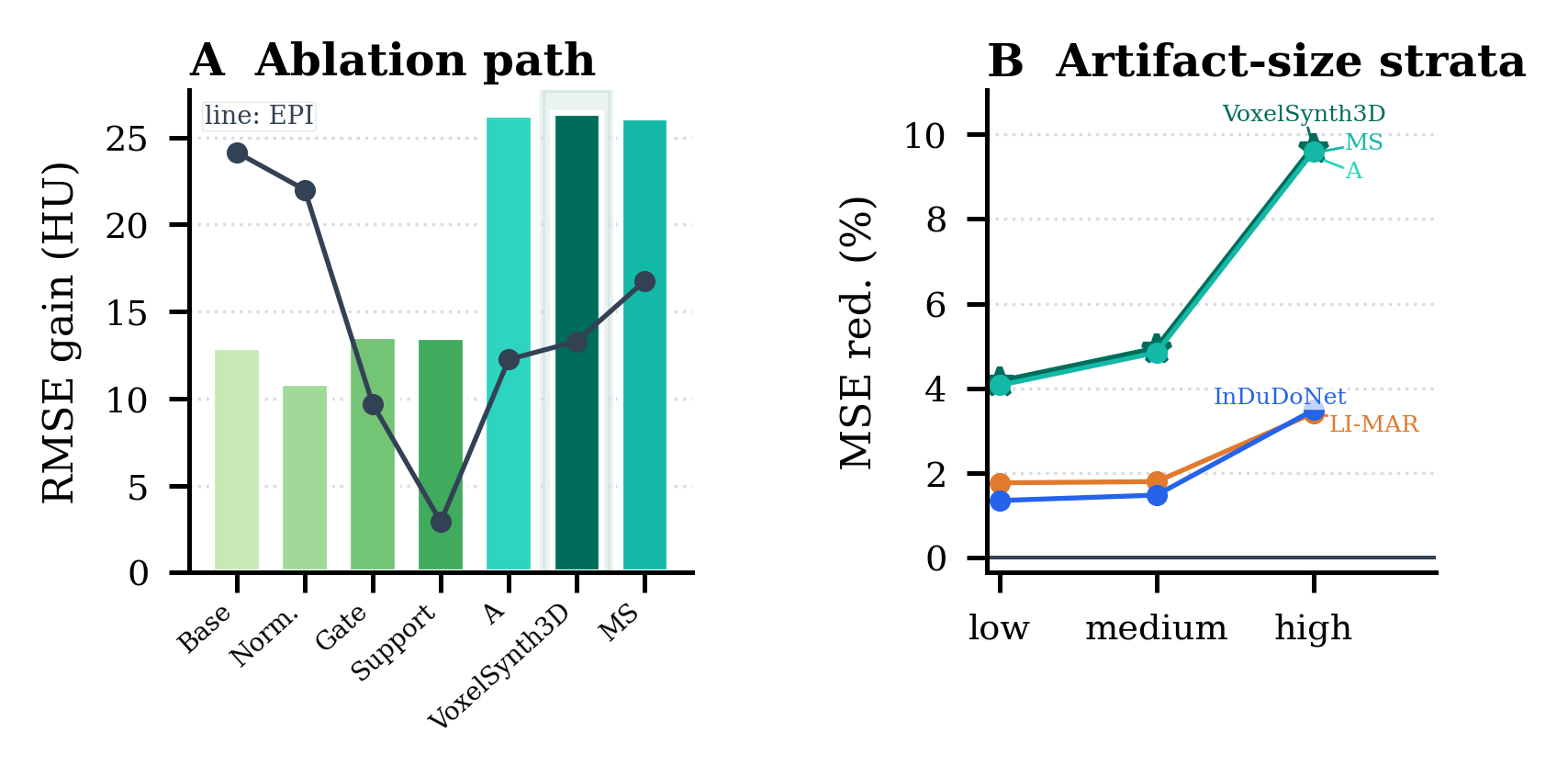}
\caption{Synthetic supporting panels. (A) Component ablation path: full-ROI RMSE gain over input with EPI overlaid. (B) Full-ROI severity by artifact size.}
\label{fig:synthetic_ablation_severity}
\end{figure}

The ablation and severity panels sharpen interpretation. Figure~\ref{fig:synthetic_ablation_severity}A shows that most of the observed full-ROI improvement comes from the volumetric prior, deviation gating, and adaptive support; edge-guided refinement is a small endpoint improvement that keeps the method training-free. VoxelSynth3D-MS improves full-ROI EPI (0.559 versus 0.542) but slightly worsens RMSE (785.72 versus 785.43 HU) and nearly doubles runtime, so it remains an ablation. Figure~\ref{fig:synthetic_ablation_severity}B shows the strongest full-ROI MSE reduction in high artifact-size cases (9.67\% versus 3.42\% for pseudo LI-MAR and 3.49\% for InDuDoNet). \rev{For continuity, these metal-containing analyses remain secondary sensitivity evidence.}

\rev{The metal-excluded spatial analysis refines interpretation near the implant. Across 40 cases at 2--5 mm, mean RMSE fell from 858.90 to 244.84 HU while mean clean-edge F1 fell from 0.548 to 0.352; paired change was $-0.197$ [95\% CI $-0.253,-0.142$]. Precision/recall 0.599/0.278 shows that retained edges were usually genuine, but many true edges were missed. Two complete misses were scored as zero F1. At 5--10/10--20 mm, F1 exceeded input by 0.089 [0.011,0.162]/0.132 [0.096,0.170]. Edge loss is therefore concentrated where correction is strongest. Lower RMSE carries a localized structural cost and does not establish recovery. Edge guidance partly improved over VoxelSynth3D-A (+0.017 EPI, +0.012 SSIM) at 1.61$\times$ runtime. \vs{} is the quality-oriented operating point; VoxelSynth3D-A is the lower-runtime option, while the validation minimum changed RMSE by only 0.09 HU.}

\subsection{\texorpdfstring{\rev{Mask Dependence and Collateral Modification}}{Mask Dependence and Collateral Modification}}
\rev{Table~\ref{tab:mask_analysis} separates correction tolerance from localization accuracy. Controlled 1--2-mm perturbations remained below the 801.48-HU input endpoint, indicating tolerance to modest boundary error. Automatic HU localization was much broader (precision 0.285): it improved target-ROI RMSE to 767.99 HU by modifying substantially more unaffected tissue.}

\begin{table}[t]
\centering
\caption{\rev{Mask sensitivity and collateral modification. Outside $\Delta$RMSE is prediction minus input RMSE in body tissue outside the fixed artifact ROI.}}
\label{tab:mask_analysis}
\scriptsize
\setlength{\tabcolsep}{2.3pt}
\renewcommand{\arraystretch}{1.02}
\begin{tabular}{@{}lrrrr@{}}
\toprule
Mask & Dice & Tissue RMSE & Outside $|\Delta|$ & Outside $\Delta$RMSE \\
 & & (HU) & (HU) & (HU) \\
\midrule
Erode 2 mm & 0.394 & 792.14 & 0.09 & +2.92 \\
Erode 1 mm & 0.749 & 788.45 & 0.15 & +4.55 \\
Exact-mask oracle & 1.000 & 786.18 & 0.15 & +4.62 \\
Dilate 1 mm & 0.814 & 784.13 & 0.14 & +4.22 \\
Dilate 2 mm & 0.609 & 780.80 & 0.18 & +4.91 \\
Automatic HU loc. & 0.380 & 767.99 & 2.97 & +39.22 \\
\bottomrule
\end{tabular}
\end{table}

\rev{For the exact-mask oracle, only 0.082\% of outside-ROI voxels changed by more than 50 HU, versus 1.04\% with automatic HU localization; outside-ROI RMSE rose by 4.62 and 39.22 HU, respectively. A larger correction region can therefore improve target RMSE while damaging normal tissue outside it. This independently shows why artifact-ROI RMSE is insufficient alone; we interpret it with clean-edge preservation and outside-ROI change. Boundary perturbations support correction-stage tolerance, not end-to-end localization accuracy. Accurate HU localization remains a deployment challenge.}

\subsection{\texorpdfstring{\rev{Validation Stability and External/Real-Data Scope}}{Validation Stability and External/Real-Data Scope}}
The 27-setting validation sweep varied edge percentile, guided-filter sigma, and blend weight. Its numerical minimum (80, 1.5 mm, 0.65; 779.29 HU) was only 0.10 HU below the conservative reported setting (85, 1.0 mm, 0.45; 779.39 HU); on test data the difference was 0.09 HU. The same full-ROI trend persisted across metal-burden strata. The uncertainty/profile analyses showed decreasing gain-CI width with sample size and the expected RMSE--EPI/SSIM tradeoff.

\rev{The mask analysis appears in Table~\ref{tab:mask_analysis}. On the external AAPM stress test, the score changed 2.2700$\rightarrow$2.4091, but its case-level 95\% CI crossed zero, $p=0.381$, and only 12/29 images improved. Repeating each 2D image into five identical slices makes this a reconstructed-image stress test, not native volumetric, leaderboard, or clinical evidence.}

On 75 real CLINIC-Metal cases, no clean target is available. \rev{Input-relative MAC change, inter-slice consistency, and runtime characterize how strongly and smoothly a method changes a volume, not whether it is anatomically correct; smoothing can improve these indicators. The cases therefore establish qualitative plausibility only, and reader- and task-based validation is required for diagnostic claims.}

\subsection{Qualitative Assessment}
\begin{figure}[ht]
\centering
    \includegraphics[width=\columnwidth]{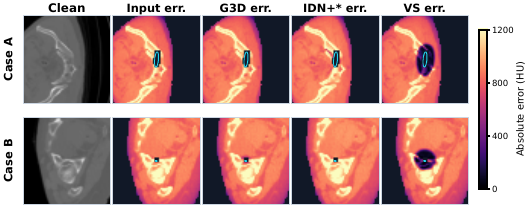}
\caption{\rev{Paired synthetic qualitative examples. Clean targets and metal-excluded absolute-error maps (0--1200 HU) are shown; darker colors indicate lower error, and cyan outlines the implant. Cases were selected by held-out gain over the shown comparators.}}
\label{fig:qualitative_synthetic}
\end{figure}

\begin{figure}[ht]
\centering
\includegraphics[width=\columnwidth]{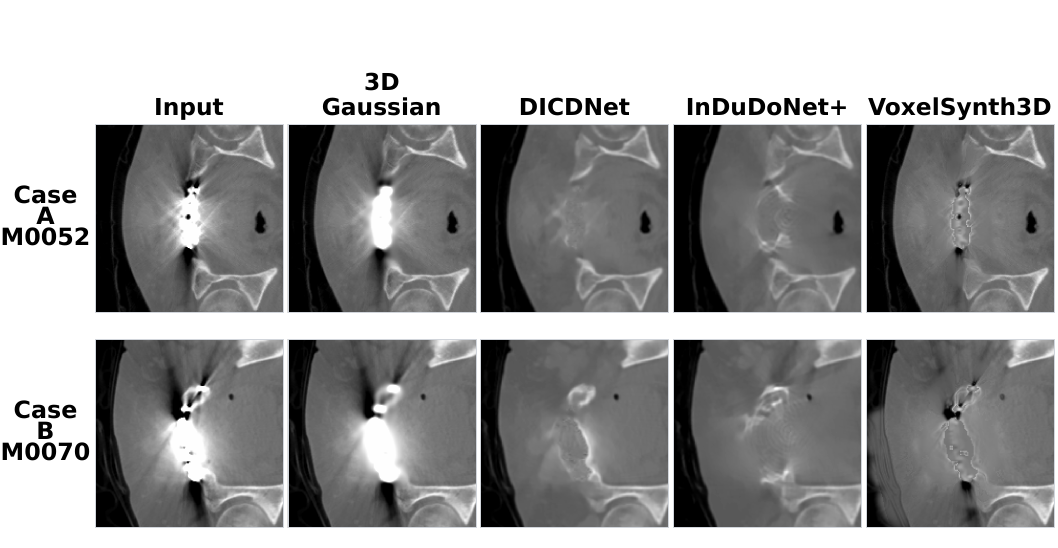}
\caption{\rev{Real CLINIC-Metal qualitative comparison (window $-900$ to 1300 HU). Paired clean targets are unavailable, so these outputs provide qualitative/no-reference evidence only.}}
\label{fig:qualitative_real}
\end{figure}

\rev{Figure~\ref{fig:qualitative_synthetic} shows two objectively selected held-out cases with large tissue-RMSE gains over the shown comparators. Darker peri-implant regions in the \vs{} error maps indicate lower absolute error under the metal-excluded endpoint used in Table~\ref{tab:main_results}. Figure~\ref{fig:qualitative_real} retains the real-case comparison as qualitative/no-reference evidence because paired targets are unavailable.}
\subsection{Limitations}
Synthetic artifacts cannot capture every scanner, protocol, implant, and reconstruction-kernel effect, and this pelvis-focused \rev{evaluation resource} does not establish generalization to other anatomies or implant geometries. \rev{The primary model supports controlled comparison; spacing-aware sensitivity retained broad-region gains but showed reduced near-metal robustness. The exact-mask oracle isolates correction; automatic HU localization over-includes high-HU anatomy.} Adapted pseudo LI/NMAR and learned checkpoints may not represent their strongest scanner-native or in-domain implementations. Finally, \vs{} lowers artifact-ROI error at the cost of lower \rev{full-ROI} EPI than some comparators, motivating better separation of anatomical edges from residual streaks. Blinded radiologist scoring remains the key clinical validation step; \rev{AAPM and real input-relative metrics cannot establish diagnostic benefit.}

\section{Conclusion}
We presented \vs, an interpretable 3D image-domain MAR framework for reconstructed CT without raw projections, with \rev{Synthetic CLINIC-Metal as a controlled paired synthetic evaluation resource. With the exact-mask oracle, it consistently reduced surrounding-tissue error versus input and the evaluated Gaussian smoother. Its support, prior, and weights are inspectable; outside-support voxels remain unchanged; targeted audits measure its edge tradeoff and localization limitations. Its limited robustness near metal under the spacing-corrected artifact model bounds this result.} Real CLINIC-metal evidence remains qualitative/no-reference. \vs{} is therefore \rev{a training-free, inspectable reconstructed-volume correction framework with measurable within-distribution benefits and limitations}, not a replacement for scanner-native MAR. \rev{Calibrated simulation and blinded task assessment are needed for external and clinical effectiveness.}

\bibliographystyle{ieeetr}
\bibliography{references}

\end{document}